\documentclass[11pt]{article}
\usepackage[final]{acl}

\usepackage{acl} 
\usepackage{times}
\usepackage{latexsym}
\usepackage[T1]{fontenc}
\usepackage[utf8]{inputenc}
\usepackage{microtype}
\usepackage{inconsolata}
\usepackage{amsmath}
\usepackage{amssymb}
\usepackage{graphicx}
\usepackage{xspace}
\usepackage{booktabs}
\usepackage{multirow}
\usepackage{lipsum}
\usepackage{subcaption}
\usepackage{threeparttable} 
\usepackage{fontawesome5}
\usepackage{hyperref}
\usepackage[table]{xcolor} 
\definecolor{rowgray}{gray}{0.92} 
\definecolor{gray_bg}{gray}{0.9} 

\usepackage{geometry}
\usepackage{enumitem} 

\usepackage{tcolorbox}
\tcbuselibrary{skins, breakable} 

\newcommand{\ours}{CoM\xspace}
 
\newcommand{\PromptRule}{\vspace{2mm}\tcbline\vspace{2mm}}

\newtcolorbox{UnifiedPromptBox}[1]{
    enhanced,
    colback=white,
    colframe=black!80,
    colbacktitle=black!80,
    coltitle=white,
    fonttitle=\bfseries\small,
    title={#1},
    sharp corners,
    boxrule=0.8pt,
    left=3mm, right=3mm, top=3mm, bottom=3mm,
    fontupper=\small,
    before skip=10pt,
    after skip=10pt,
    breakable,   
}

\newtcolorbox{TemplateBlock}{
    colback=gray!10,
    colframe=gray!10,
    sharp corners,
    boxrule=0pt,
    left=2mm, right=2mm, top=2mm, bottom=2mm,
    fontupper=\ttfamily\small,
    before skip=2mm,
    after skip=2mm
}

\title{Chain-of-Memory: Lightweight Memory Construction with \\ Dynamic Evolution for LLM Agents}

\author{
    Xiucheng Xu$^{1,2,3}$, 
    Bingbing Xu$^{1,2}$\thanks{\; Corresponding author.}, 
    Xueyun Tian$^{1,2,3}$, 
    \textbf{Zihe Huang}$^{1,2,3}$,\\
    \textbf{Rongxin Chen}$^{1,2,3}$,
    \textbf{Yunfan Li}$^{1,2,3}$, 
    \textbf{Huawei Shen}$^{1,2,3}$\\
        $^{1}$State Key Laboratory of AI Safety, Beijing, 100086 \\
        $^{2}$Institute of Computing Technology, Chinese Academy of Sciences \\
        $^{3}$University of Chinese Academy of Sciences \\
        \texttt{\{xuxiucheng24s,xubingbing\}@ict.ac.cn}
}

\begin{document}
\maketitle

\begin{abstract}

External memory systems are pivotal for enabling Large Language Model (LLM) agents to maintain persistent knowledge and perform long-horizon decision-making. Existing paradigms typically follow a two-stage process: computationally expensive memory construction (e.g., structuring data into graphs) followed by naive retrieval-augmented generation. However, our empirical analysis reveals two fundamental limitations: complex construction incurs high costs with marginal performance gains, and simple context concatenation fails to bridge the gap between retrieval recall and reasoning accuracy. To address above challenges, we propose \textbf{CoM (Chain-of-Memory)}, a novel framework that advocates for a paradigm shift toward lightweight construction paired with sophisticated utilization. \ours introduces a \textit{Chain-of-Memory} mechanism that organizes retrieved fragments into coherent inference paths through dynamic evolution, utilizing adaptive truncation to prune irrelevant noise. Extensive experiments on the LongMemEval and LoCoMo benchmarks demonstrate that \ours outperforms strong baselines with accuracy gains of 7.5\%--10.4\%, while drastically reducing computational overhead to approximately 2.7\% of token consumption and 6.0\% of latency compared to complex memory architectures.
Code is available at \url{https://github.com/Xiucheng-Xu/CoM}.

\end{abstract}

\section{Introduction}

Large Language Model (LLM)-driven agents are evolving from simple conversational interfaces into autonomous entities capable of handling complex, long-horizon tasks~\cite{yang2024swe, singh2025agentic, zhang2025landscape}. Effective decision-making in such scenarios necessitates the continuous integration of extensive interaction histories. However, the inherent constraints of finite context windows~\cite{fei2024extending, zhang2025survey} in LLMs reduce their ability to retain vast amounts of information, which limits their potential for long-term knowledge accumulation and adaptability~\cite{wang2024adapting, liu2024lost}.
As a result, equipping agents with explicit external memory systems~\cite{zhong2024memorybank, li2025memos} has emerged as a critical component for enabling persistent knowledge accumulation and long-horizon decision-making in LLM-based agents.

Existing works on agent memory generally adopt a two-stage paradigm: memory construction followed by memory utilization. During construction, raw interaction traces are often transformed into structured formats, such as trees or graphs, to capture semantic connections~\cite{chhikara2025mem0, xu2025mem}, resulting in high computational costs. Conversely, memory utilization typically adopts naive Retrieval-Augmented Generation (RAG)~\cite{gao2023retrieval}, a conventional retrieve-and-concatenate paradigm where retrieved fragments are directly incorporated into prompts.

\begin{figure*}[t] 
    \centering
    \includegraphics[width=\textwidth]{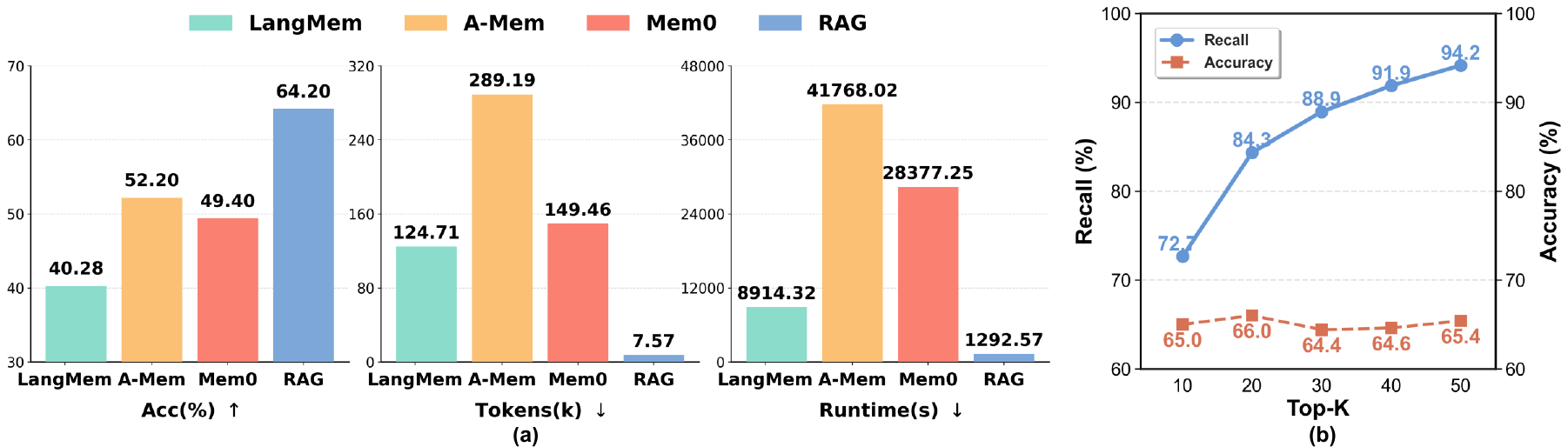}
    
    \caption{\textbf{Empirical limitations of existing paradigms.} (a) Heavy-weight memory construction strategies fail to demonstrate cost-effectiveness. (b) Naive retrieval strategies exhibit a reasoning bottleneck, where retrieved evidence is not effectively utilized for answer generation.}
    \label{fig:intro_fig}
\end{figure*}

However, as Fig.~\ref{fig:intro_fig} illustrated, we find that the above methods suffer from two fundamental limitations:
First, our empirical observations reveal that the heavy cost incurred by elaborate memory construction is not matched by commensurate performance gains. Specifically, on long-term memory QA benchmarks, such structured memories lead to significantly higher token usage and latency, while yielding negligible accuracy improvements over a naive RAG baseline.
Second, there exists a clear gap between the retrieval effectiveness and final response accuracy: even when ground-truth evidence is successfully retrieved, directly injecting retrieved fragments into the prompt often fails to translate recall into accurate answers, suggesting that 
the prevailing practice of merely inserting fragments into prompts is insufficient for reasoning.

Motivated by these observations, we argue for a paradigm shift in memory design: rather than investing in costly and elaborate memory construction coupled with naive retrieve-and-concatenate utilization, memory systems should adopt lightweight construction with more principled and effective utilization.
This shift requires rethinking how retrieved memories are organized and consumed during generation. In particular, we identify two key limitations in existing memory utilization strategies. First, retrieved memories are processed largely in isolation, without explicit relational structures to support compositional or multi-hop reasoning. Second, relevant evidence is often embedded within long and noisy contexts, where excessive and irrelevant information obscures critical signals, making it difficult for the model to effectively leverage retrieved memories.


Inspired by these, we propose a Chain-of-Memory framework, named \ours, featuring lightweight memory construction for retrieval and a dynamic memory chain evolution mechanism for organizing isolated fragments into coherent inference paths. Specifically, we construct these memory chains by jointly evaluating query relevance and contextual consistency. Furthermore, to mitigate information redundancy, we adopt an adaptive truncation mechanism that prunes non-essential contexts, preventing critical information from being diluted.
Extensive experiments on the LongMemEval~\cite{wu2024longmemeval} and LoCoMo~\cite{maharana2024evaluating} benchmarks demonstrate that \ours consistently outperforms the strongest baselines, yielding absolute accuracy gains of \textbf{7.5\%}--\textbf{10.0\%} with GPT-4o-mini and \textbf{9.1\%}--\textbf{10.4\%} with Qwen3-32B.
Meanwhile, \ours reduces the computational footprint drastically, requiring approximately \textbf{2.7\%} of the token consumption and merely \textbf{6.0\%} of the time compared to prevailing complex memory architectures.



Our contributions can be summarized as follows:
\begin{itemize}
    \item \textbf{Paradigm-Shifting:} We advocate for a fundamental shift away from costly, structured memory construction coupled with naive retrieval, towards a principle of lightweight construction paired with systematic and dynamic memory utilization.

    \item \textbf{Methodologically Novel:} We propose the Chain-of-Memory framework, which organizes retrieved fragments into coherent inference paths through dynamic memory chain evolution and prunes non-essential contexts via adaptive truncation.

    \item \textbf{Empirically Effective:} The framework delivers substantial accuracy gains (7.5\%–10.4\%) while drastically reducing computational costs (to ~2.7\% token usage and ~6.0\% runtime) compared to existing complex memory structures.

\end{itemize}

\section{Related Work}

\subsection{Retrieval-Augmented Generation}

Standard Retrieval-Augmented Generation (RAG) systems~\cite{gao2023retrieval, yu2024chain} partition external corpora into discrete segments, retrieving relevant chunks based on semantic similarity to augment LLM prompts.
Prevailing chunking strategies include rule-based methods that generate fixed-size segments~\cite{sarthi2024raptor, liu2025passage}, semantic-based approaches that cluster content by topic~\cite{qu2025semantic}, and LLM-driven techniques that leverage parametric knowledge for segmentation~\cite{pan2025secom, duarte2024lumberchunker, zhao2024meta}.
Some approaches adopt GraphRAG systems~\citep{guo2024lightrag, dong2025youtu} to enhance recall through extensive relationship pre-computation.
However, these frameworks operate on static data and lack mechanisms to maintain an evolving memory of historical interactions. In contrast, agent memory systems are grounded in environmental interaction, dynamically assimilating information from agent actions and environmental feedback into a persistent memory store~\cite{wang2025inducing, zhao2024expel, sun2025rearter}.

\subsection{Memory for LLM Agents} 

Memory systems enable LLM-driven agents to maintain persistent context and perform consistent decision-making in complex environments~\cite{liu2025advances, mei2025survey}.
Early approaches typically organize historical experiences as linear sequences, occasionally augmented with hierarchical structures.
For instance, MemGPT~\cite{packer2023memgpt} draws inspiration from operating system virtual memory to manage context paging, while frameworks like SCM~\cite{wang2023enhancing} employ controller-based mechanisms to optimize information retention.
While implementationally efficient, these linear methods often struggle to capture explicit semantic dependencies between temporally distant fragments. To address this, some approaches organize memory into complex structures, such as trees or graphs. Systems such as MemTree~\cite{rezazadeh2024isolated}, Zep~\cite{rasmussen2025zep}, A-Mem ~\cite{xu2025mem}, Mem0~\cite{chhikara2025mem0} construct temporal knowledge trees, graphs or networks to permit dynamic updates and relational retrieval.

Although these structured approaches significantly improve reasoning consistency, they typically incur high computational costs during the memory construction phase, making them less practical for real-time applications.
Furthermore, despite the sophisticated organization, the utilization stage frequently relies on naive retrieval paradigms that fail to support compositional reasoning. Diverging from these heavy-construction frameworks, our \ours proposes a paradigm shift: minimizing construction overhead while enhancing utilization through a dynamic chain of memory evolution chain-of-evidence mechanism.

\section{Methodology}
\label{sec:methodology}

\begin{figure*}[t] 
    \centering
    \includegraphics[width=1\textwidth]{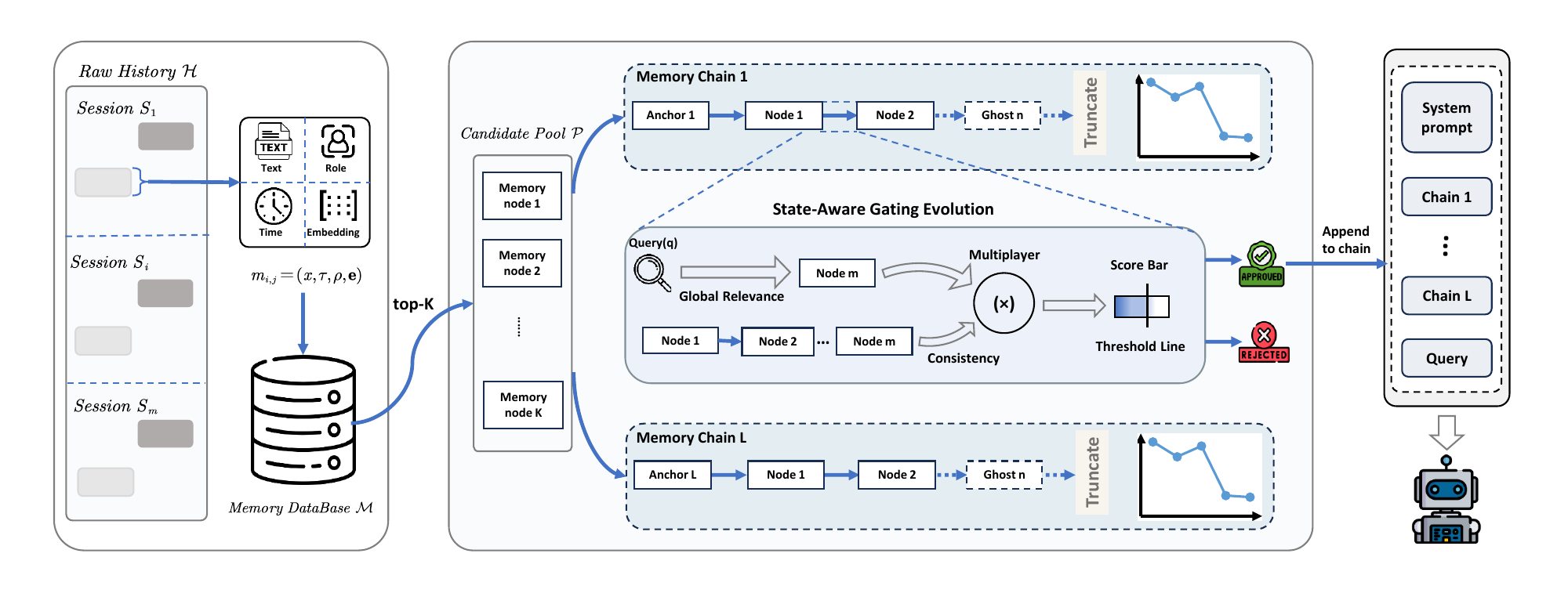} 
    \caption{\textbf{The overview architecture of \ours.} The workflow consists of two stages: (1) Memory Construction and Retrieval, and (2) Dynamic Memory Chain Evolution. The dashed Ghost n denotes a candidate node whose gating score sharply drops below the adaptive threshold, thereby triggering path truncation.}
    \label{fig:main} 
\end{figure*}

We propose \ours, a framework that integrates a lightweight flat index with a dynamic post-retrieval chaining mechanism. Instead of relying on complex pre-built structures, \ours constructs query-oriented dependencies among memory nodes to synthesize coherent reasoning paths. As illustrated in Figure~\ref{fig:main}, the framework comprises two core stages: (1) Memory Construction and Retrieval, and (2) Dynamic Memory Chain Evolution.

\subsection{Memory Construction and Retrieval}
To minimize computational overhead and temporal latency, we adopt a lightweight, flat memory architecture that stores raw conversation data without complex pre-structuring. By leveraging direct semantic similarity for retrieval, this streamlined approach ensures rapid access to relevant context, avoiding the heavy computational burden associated with hierarchical or graph-based modeling.

\paragraph{Construction.}
We formalize the raw conversation history as $\mathcal{H} = \{S_1, \dots, S_M\}$. Each session $S_i = (t_{i,1}, \dots, t_{i,N_i})$ consists of a chronological sequence of $N_i$ turns, where the turn $t_{i,j}$ is a single utterance generated by either the user or the assistant. To facilitate efficient indexing, we treat a single turn as the atomic granularity for memory construction . Specifically, each turn $t_{i,j}$ is transformed into a Memory Node:
\begin{equation}
    m_{i,j} = (x, \tau, \rho, \mathbf{e}),
\end{equation}
where $x$ denotes the raw textual content, $\tau$ is the timestamp, $\rho \in \{\text{User}, \text{Assistant}\}$ indicates the speaker role, and $\mathbf{e} \in \mathbb{R}^d$ is the embedding of $x$. We define the resulting memory database as $\mathcal{M} = \{ (m_{i,1}, \dots, m_{i,N_i}) \}_{i=1}^M$ and simplify this notation to $\mathcal{M} = \{ m_{1}, \dots, m_{n} \}$. This organization preserves the raw structure of the original conversation, supporting both context-aware filtering and efficient retrieval of atomic memory nodes.

\paragraph{Retrieval.}
For a given query $q$, we employ an embedding model to obtain its vector representation $\mathbf{q} = E(q)$, and compute the relevance score for each memory node $m_i$ via cosine similarity:
\begin{equation}
    s_{i} = \cos(\mathbf{q}, \mathbf{e}_{i}).
\end{equation}
The Top-$K$ memory nodes with the highest scores are selected to form a candidate pool $\mathcal{P} = \{m_1, m_2, \dots, m_K\} \subset \mathcal{M}$. Although $\mathcal{P}$ provides the necessary context, memory nodes within this pool remain isolated, lacking explicit logical connections to guide the reasoning process.

\subsection{Dynamic Memory Chain Evolution}
To construct coherent and contextually relevant reasoning paths from the retrieved memories, we propose a Dynamic Memory Chain Evolution mechanism. This stage iteratively expands memory chains by selecting subsequent nodes that adhere to the user's original intent, while maintaining logically contextual coherence with the ongoing reasoning path. By dynamically evolving these chains, we  model long-range dependencies while pruning irrelevant information which ensures high quality context for reasoning.

\paragraph{Initialization.}
The process commences by selecting the Top-$L$ memory nodes from the candidate pool $\mathcal{P}$ according to their retrieval scores to serve as starting anchors. Consequently, $L$ distinct memory chains are constructed, each initialized as $\mathcal{C}_z^{(0)} = \{m_z\}$. These initial anchors serve as the foundation for the subsequent iterative expansion, where each chain evolves step-by-step to incorporate complementary information.

\paragraph{State-Aware Gating Evolution.}
At expansion step $t$ for a specific chain $\mathcal{C}_z^{(t)}$, we aim to identify the optimal successor from the candidates. To strictly control chain growth quality, we compute a gating score $S_{\text{gate}}$ for each candidate $m$:
\begin{equation}
\small 
    S_{\text{gate}}(m) = \underbrace{\cos(\mathbf{m}, \mathbf{q})}_{\text{Global Relevance}} \times \underbrace{\cos(\mathbf{m}, \mathbf{C}_z^{(t)})}_{\text{Contextual Consistency}},
\end{equation}
where $\mathbf{C}_z^{(t)}$ denotes the embedding of chain $\mathcal{C}_z^{(t)}$ at step $t$. This multiplicative formulation acts as a soft logical gate, ensuring that a valid successor satisfies two essential criteria: alignment with the user's query (Global Relevance) and coherence with the current reasoning context (Contextual Consistency). Consequently, the candidate $m^*$ maximizing $S_{\text{gate}}$ is selected as the next node.

\paragraph{Adaptive Path Truncation.}
To mitigate the risk of semantic drift—where the chain diverges into irrelevant topics, we implement a relative threshold termination strategy. Let $s_{t-1}$ denote the gating score of the node appended in the previous step, and $s_t^*$ be the optimal score among candidates in the current step. The expansion terminates if the relevance exhibits a sharp attenuation:
\begin{equation}
    s_t^* < \beta \cdot s_{t-1},
\end{equation}
where $\beta \in [0, 1]$ is a hyperparameter controlling the sensitivity of the truncation. This condition identifies "cliff-like" drops in semantic relevance, indicating that further expansion would yield diminishing returns. Thus, we finalize the chain at this point to preserve the integrity and quality of the reasoning path.

\subsection{Complexity Analysis}
\label{sec:complexity}

We analyze the complexity of the dynamic memory chain evolution process. Let $K$ be the size of the retrieved candidate pool, $L$ the number of initialized memory chains, and $d$ the embedding dimension. Within each chain, selected candidates are removed from that chain's candidate set, while different chains may still reuse the same candidates. If adaptive truncation is never triggered, a single chain evaluates at most $\sum_{t=1}^{K-1}(K-t)$ candidates, resulting in $O(K^2d)$ vector similarity operations. Across $L$ independently evolved chains, the worst-case complexity is therefore $O(LK^2d)$. If a maximum chain length $T$ is imposed, the bound becomes $O(LTKd)$, where $T \le K$.

In practice, both $L$ and $K$ are deliberately kept small, and adaptive truncation usually stops chain expansion before the worst case is reached. Moreover, query-candidate relevance scores can be precomputed once for the retrieved pool, so the main repeated computation comes from contextual consistency scoring during chain evolution. As shown by the end-to-end runtime results, this overhead remains close to that of naive RAG and is substantially lower than methods that require expensive graph or tree memory construction.

\section{Experiments}
\label{sec:experiments}

\subsection{Experimental Setup}
\label{sec:exp_setup}

\paragraph{Datasets.} 

\begin{table}[t]
    \small
    \centering
    
    \resizebox{\linewidth}{!}{
        \begin{tabular}{llrr}
            \toprule
            \multirow{2}{*}{\textbf{Dataset}} & \multirow{2}{*}{\textbf{Category}} & \multicolumn{2}{c}{\textbf{\# Questions}} \\
            \cmidrule(lr){3-4}
             & & \textbf{Count} & \textbf{Ratio} \\
            \midrule
            
            \multirow{4}{*}{LongMemEval} 
             & Single-hop (S-hop) & 156 & 31.2\% \\
             & Multi-hop (M-hop) & 133 & 26.6\% \\
             & Temporal (Temp) & 133 & 26.6\% \\
             & Knowledge (Kno) & 78 & 15.6\% \\
            \midrule
            
            \multirow{4}{*}{LoCoMo} 
             & Single-hop (S-hop) & 841 & 54.6\% \\
             & Multi-hop (M-hop) & 282 & 18.3\% \\
             & Temporal (Temp) & 321 & 20.8\% \\
             & Knowledge (Kno) & 96 & 6.2\% \\
            \bottomrule
        \end{tabular}
    }
    
    \vspace{2pt}
    \begin{minipage}{\linewidth} 
        \scriptsize
        \textit{Note:} LongMemEval `Single-hop' aggregates all single-session subtypes. LoCoMo excludes 446 `Adversarial' samples.
    \end{minipage}

    \caption{Distribution of unified question categories for LongMemEval and LoCoMo.}
    \label{tab:unified_stats_final}
\end{table}

We evaluate our method on two long-term memory benchmarks: LongMemEval (specifically the LongMemEval-S split)~\cite{wu2024longmemeval} and LoCoMo~\cite{maharana2024evaluating}. 
LongMemEval consists of 500 QA pairs derived from 500 distinct conversation histories, with an average context length of approximately 115k tokens. 
LoCoMo contains 1,986 QA instances from 10 conversation sets, averaging approximately 26k tokens per context.
To maintain consistency across benchmarks, we map the diverse original question types into four unified categories: \textit{single-hop}, \textit{multi-hop}, \textit{temporal reasoning}, and \textit{knowledge}. Following Mem0~\cite{chhikara2025mem0}, we exclude the \textit{adversarial} category from LoCoMo to ensure fair comparison with baselines.
Table~\ref{tab:unified_stats_final} presents the statistics of these unified categories. 

\paragraph{Baselines.} 
We compare our approach against the following baselines: 
(1) \textbf{Full Context}, which inputs the complete conversation history directly; 
(2) \textbf{Naive RAG}~\cite{gao2023retrieval}, evaluated with both \textit{session-level} and \textit{turn-level} retrieval granularities; 
(3) \textbf{LangMem}~\cite{Chase_LangChain_2022}, which extracts salient facts for vector-based retrieval; 
(4) \textbf{A-Mem}~\cite{xu2025mem}, which builds knowledge graphs with dynamic note linking; and 
(5) \textbf{Mem0}~\cite{chhikara2025mem0}, which utilizes incremental fact compression with graph extensions.

\paragraph{Metrics.} 
We evaluate performance based on both effectiveness and efficiency. 
For effectiveness, we report \textbf{Accuracy (ACC)}, utilizing LLM-as-LLM-Judge to determine if the response correctly answers the question. 
Unlike surface-level metrics such as F1 or BLEU which rely on lexical overlap, this approach better captures semantic alignment. 
For efficiency, we report Token Consumption and Total Runtime as end-to-end metrics, encompassing memory construction, retrieval, and LLM generation.
We include these metrics because many existing memory architectures prioritize performance over efficiency, often overlooking the significant computational overhead and latency associated with building and maintaining complex memory structures. 
Such neglect limits their practical deployment in real-world scenarios.

\paragraph{Implementation Details.}
We employ GPT-4o-mini~\cite{hurst2024gpt} and Qwen3-32B (Non-Thinking Mode)~\cite{qwen3technicalreport} as the backbone Large Language Models (LLMs) for both answer generation and the LLM-as-Judge evaluation.For semantic representation, we utilize Qwen3-Embedding-8B~\citep{qwen3embedding} to encode memory nodes and queries. Regarding retrieval, the number of retrieved segments ($k$) is set to 20 across all datasets. To ensure a fair comparison, we standardize the answer generation prompts for all experiments. Detailed prompts for generation and judge are provided in Appendix~\ref{sec:appendix_prompts}

\subsection{Main Results}
\label{sec:main_results}

\begin{table*}[t]
\centering
\resizebox{\textwidth}{!}{
    \setlength{\tabcolsep}{3.2pt} 
    \renewcommand{\arraystretch}{1.15} 
    
    \begin{tabular}{l ccccc rc @{\hskip 0.2in} ccccc rc}
    \toprule
    
    \multirow{3}{*}{\textbf{Method}} & 
    \multicolumn{7}{c}{\textbf{LongMemEval}} & 
    \multicolumn{7}{c}{\textbf{LoCoMo}} \\
    \cmidrule(lr){2-6} \cmidrule(l r{16pt}){7-8} \cmidrule(l{2pt} r){9-13} \cmidrule(l r{2pt}){14-15}
    
    & \multicolumn{5}{c}{Accuracy (\%) $\uparrow$} & \multicolumn{2}{c}{Cost $\downarrow$} 
    & \multicolumn{5}{c}{Accuracy (\%) $\uparrow$} & \multicolumn{2}{c}{Cost $\downarrow$} \\

    \cmidrule(lr){2-6} \cmidrule(l r{16pt}){7-8}  \cmidrule(l{2pt} r){9-13} \cmidrule(l r{2pt}){14-15}
    
    & S-hop & M-hop & Temp & Kno & Total & Token & Time 
    & S-hop & M-hop & Temp & Kno & Total & Token & Time \\
    \midrule
    
    \multicolumn{15}{l}{\textit{\textbf{Backbone: GPT-4o-mini}}} \\
    \addlinespace[2pt] 
    
    Full-Context  & 68.59 & 39.85 & 44.36 & 76.92 & 55.80 & 112.7 & 8154 & \textbf{88.59} & \textbf{53.55} & 50.78 & 50.00 & 71.88 & 31.8 & 4933 \\
    RAG (session) & 75.64 & 54.89 & 44.36 & 57.69 & 59.00 & 26.6 & 1512 & 78.83 & 52.02 & 47.16 & 44.79 & 65.32 & 7.5 & 2601 \\
    RAG (turn)    & 79.49 & 54.14 & 48.87 & 76.92 & 64.20 & \textbf{7.6} & \textbf{1293} & 77.76 & 41.13 & 44.79 & \textbf{60.44} & 65.39 & \textbf{1.4} & \textbf{2304} \\
    LangMem       & 49.35 & 38.35 & 33.83 & 35.89 & 40.20 & 124.7 & 8914 & 66.46 & 36.17 & 41.74 & 38.54 & 54.02 & 286.8 & 3318 \\
    A-Mem         & 73.72 & 42.86 & 34.59 & 55.13 & 52.20 & 289.2 & 41768 & 74.31 & 37.58 & 44.54 & 37.50 & 59.09 & 357.0 & 8787 \\
    Mem0          & 41.67 & 45.11 & 54.14 & 64.10 & 49.40 & 149.5 & 28377 & 73.72 & 40.07 & 61.05 & 47.91 & 63.31 & 388.4 & 6326 \\
    \textbf{\ours} & \textbf{84.62} & \textbf{65.41} & \textbf{65.41} & \textbf{83.33} & \textbf{74.20} & 8.2 & 2730 & 83.59 & 48.94 & \textbf{73.52} & 46.88 & \textbf{72.86} & 4.4 & 3058 \\
    
    \midrule 
    \addlinespace[4pt] 
    
    \multicolumn{15}{l}{\textit{\textbf{Backbone: Qwen3-32B}}} \\
    \addlinespace[2pt]
    
    Full-Context  & 73.72 & 41.35 & 38.35 & 70.51 & 55.20 & 119.6 & 11504 & \textbf{86.44} & \textbf{51.06} & 49.53 & 45.83 & 69.74 & 36.0 & 2981 \\
    RAG (session) & 78.21 & 54.89 & 47.37 & 66.67 & 62.00 & 27.3 & \textbf{1598} & 78.83 & 45.39 & 44.85 & 39.58 & 63.18 & 7.8 & \textbf{1816} \\
    RAG (turn)    & 82.05 & 59.40 & 48.87 & 74.36 & 66.00 & \textbf{7.9} & 2891 & 75.26 & 34.04 & 57.32 & 41.66 & 61.88 & \textbf{1.7} & 2604 \\
    LangMem       & 51.28 & 39.10 & 35.33 & 37.17 & 41.60 & 150.2 & 7635 & 67.89 & 38.65 & 42.05 & 37.50 & 55.25 & 336.6 & 3850 \\
    A-Mem         & 78.20 & 39.10 & 30.08 & 66.67 & 53.20 & 331.8 & 32949 & 77.88 & 41.48 & 40.49 & 36.45 & 60.84 & 409.7 & 9721 \\
    Mem0          & 54.49 & 45.86 & 43.61 & 62.82 & 50.60 & 156.0 & 19721 & 75.74 & 45.74 & 53.27 & 42.70 & 63.51 & 456.0 & 8739 \\
    \textbf{\ours} & \textbf{86.54} & \textbf{69.92} & \textbf{69.17} & \textbf{79.49} & \textbf{76.40} & 8.8 & 2002 & 81.45 & 46.09 & \textbf{71.96} & \textbf{48.95} & \textbf{70.97} & 4.8 & 2050 \\
    
    \bottomrule
    \end{tabular}
}
\caption{Main results on the LongMemEval and LoCoMo benchmarks. We report Accuracy (\%) across four sub-tasks: \textbf{S-hop} (Single-hop), \textbf{M-hop} (Multi-hop), \textbf{Temp} (Temporal), and \textbf{Kno} (Knowledge). \textbf{Total} denotes the accuracy calculated across the entire dataset (all samples). \textbf{Token} refers to the total end-to-end token consumption (in thousands, $k$), and \textbf{Time} indicates the total runtime in seconds ($s$). Best results are highlighted in bold.}
\label{tab:main_results}
\end{table*}

Table~\ref{tab:main_results} presents the comparative performance of \ours against various baselines on the LongMemEval and LoCoMo benchmarks.
\ours demonstrates a superior trade-off between effectiveness and efficiency, achieving state-of-the-art accuracy while maintaining a minimal computational cost.

\paragraph{Performance and Effectiveness.}
Our method demonstrates significant improvements over existing state-of-the-art baselines.
On the \textbf{LongMemEval} benchmark, \ours secures the highest total accuracy. With the Qwen3-32B backbone, it achieves \textbf{76.40\%}, exceeding the strongest baseline by approximately \textbf{15.8\%}.
These gains are particularly pronounced in logic-intensive sub-tasks, such as \textit{Multi-hop} and \textit{Temporal Reasoning}, where our method effectively bridges information gaps that limit other approaches.
Similarly, on the \textbf{LoCoMo} benchmark, \ours continues to lead, outperforming the best baseline by over \textbf{14\%} in total accuracy (70.97\% vs. 61.88\% for Qwen3-32B).
Notably, in the challenging \textit{Temporal} task, our method shows a remarkable advantage, improving accuracy by nearly \textbf{25\%} compared to the closest competitor.
These consistent improvements show our dynamic chaining mechanism successfully organizes isolated fragments into coherent inference paths, addressing the "isolated memory" limitation observed in prior works.

\paragraph{Efficiency and Cost-Benefit Analysis.}
A critical observation from Table~\ref{tab:main_results} is the inefficiency of existing complex memory organizations.
Despite their intricate designs, methods like A-Mem and Mem0 perform comparably to—or sometimes worse than—simple baselines, failing to justify their complexity.
More strikingly, these structures incur prohibitive computational costs that can exceed even the \textit{Full-Context} baseline. For example, constructing and querying A-Mem with Qwen3 requires nearly \textbf{332k} tokens—almost \textbf{3$\times$} the consumption of processing the entire full context (119.56k)—yet it yields lower accuracy on LongMemEval. In sharp contrast, \ours maintains a highly efficient profile. It operates with a minimal token footprint (approx. \textbf{8.8k}), which is comparable to the most lightweight baselines. Furthermore, regarding latency, \ours significantly reduces the total runtime compared to graph-based methods (e.g., \textbf{2,002s} vs. 32,949s for A-Mem), avoiding the overhead of complex structure traversal.

\paragraph{Comparison with re-ranking.}
We further compare CoM with a cross-encoder re-ranking baseline in Appendix~\ref{app:reranking_comparison}. The results show that stronger post-retrieval scoring alone cannot match CoM, suggesting that context-aware chain organization is essential for the observed gains.

\subsection{Ablation Study}
\label{sec:ablation_study}

\begin{figure*}[t]
    \centering
    \includegraphics[width=1.0\linewidth]{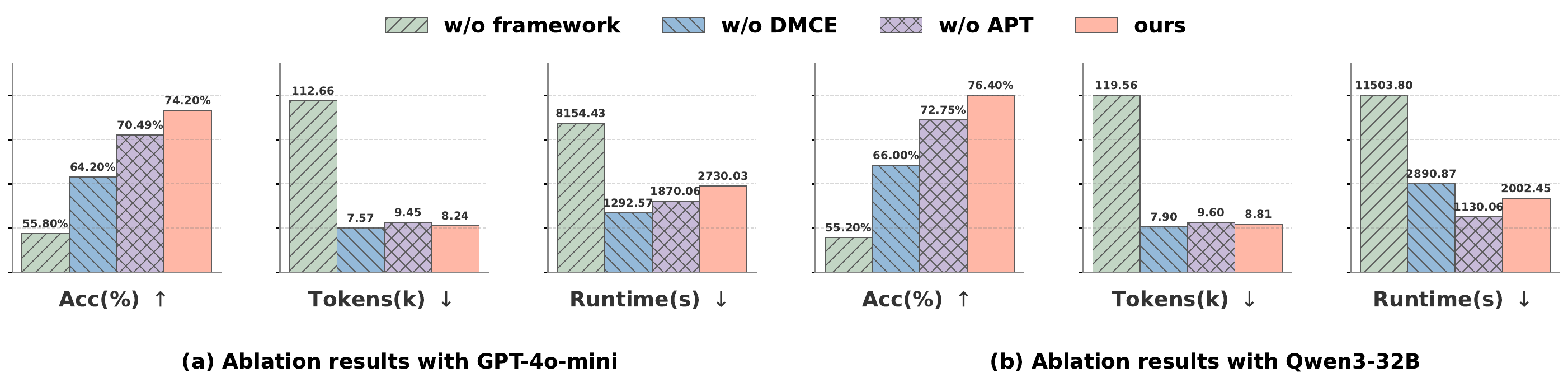}
    \caption{\textbf{Ablation Study Results.} We compare the performance of our full method against variants removing specific components (w/o Framework, w/o DMCE, w/o APT) on GPT-4o-mini (a) and Qwen3-32B (b). The metrics include Accuracy (Acc), Token consumption, and Runtime. Our method achieves the best trade-off between accuracy and efficiency.}
    \label{fig:ablation}
\end{figure*}

\paragraph{Effect of framework components.}
To rigorously assess the contribution of each component, Figure~\ref{fig:ablation} details our ablation study on Dynamic Memory Chain Evolution (DMCE) and Adaptive Path Truncation (APT). 
First, removing the entire framework causes accuracy on GPT-4o-mini to plummet to 55.80\%, accompanied by a massive $13\times$ surge in token consumption (112.66 vs. 8.24). This confirms that our architecture is foundational for maintaining long-context coherence while minimizing redundancy. 
Second, excluding DMCE leads to a significant degradation, with accuracy falling to 59.00\%. Although the \textit{w/o DMCE} variant yields marginally lower latency on Qwen3-32B, it suffers a severe drop in answer quality, proving that Chain Evolution is critical for sustaining information density without imposing heavy overhead. 
Finally, omitting the APT stage results in suboptimal accuracy (72.75\%) and increased computational costs. This outcome underscores the necessity of Path Truncation in filtering out irrelevant noise and balancing efficiency with precision.

\paragraph{Effect of the gating formulation.}
We further analyze the state-aware gating score by replacing the multiplicative formulation with three alternatives: using only global relevance, using only contextual consistency, and using a weighted average of the two scores. As shown in Table~\ref{tab:gating_ablation}, the multiplicative formulation performs best on both datasets. This supports our design choice of treating query relevance and contextual consistency as a soft conjunction: a useful successor should be relevant to the original query while also coherent with the evolving reasoning chain.

\begin{table}[t]
    \small
    \centering
    \renewcommand{\arraystretch}{1.2} 
    
    \setlength{\tabcolsep}{3pt} 

    \begin{tabular}{lcc}
        \toprule
        \textbf{Gating Variant} & \textbf{LongMemEval} & \textbf{LoCoMo} \\
        \midrule
        Only Global Relevance & 66.15 & 65.26 \\
        Only Contextual Consistency & 69.35 & 66.84 \\
        Weighted Average & 72.55 & 68.17 \\
        Multiplicative Combination & \textbf{76.40} & \textbf{70.97} \\
        \bottomrule
    \end{tabular}
    
    \caption{Ablation of the state-aware gating formulation using Qwen3-32B. Accuracy is reported in percentage.}
    \label{tab:gating_ablation}
\end{table}

\subsection{Hyperparameter Analysis}
\label{sec:hyperparameter_analysis}

\begin{figure}[t]
    \centering
    \includegraphics[width=1.0\columnwidth]{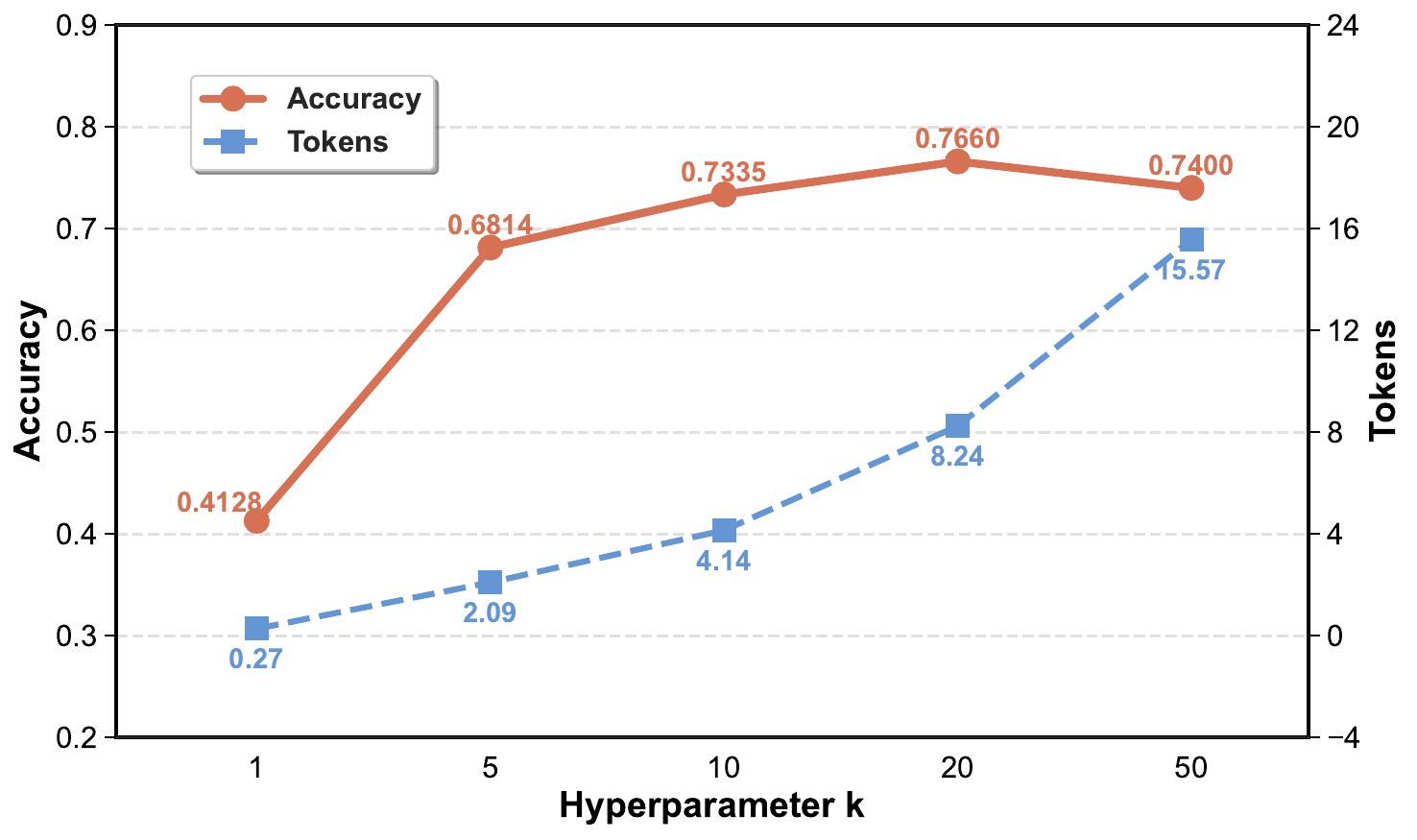}
    \caption{The impact of hyperparameter $k$ on model accuracy and computational cost (tokens).}
    \label{fig:para_k}
\end{figure}

\paragraph{Effect of retrieval size.}
We investigate the impact of the retrieval hyperparameter $k$ ($k \in \{1, 5, 10, 20, 50\}$) on LongMemEval using the Qwen3-32B backbone. 
As illustrated in Figure~\ref{fig:para_k}, the performance gains from increasing $k$ eventually saturate, suggesting that pivotal information is predominantly concentrated within the top retrieved chunks. 
Consequently, excessively increasing $k$ yields diminishing returns, as the inclusion of irrelevant context introduces noise that may hamper reasoning. 
In terms of efficiency, although token consumption grows linearly with $k$, total runtime remains largely stable because latency is dominated by LLM inference and network overhead rather than retrieval. 
Thus, a moderate $k$ provides an effective trade-off between context sufficiency and computational efficiency.

\paragraph{Sensitivity to the truncation threshold.}
We evaluate the adaptive truncation threshold $\beta$, which controls when chain expansion stops after a sharp score drop. 
Table~\ref{tab:beta_sensitivity} shows that $\beta=0.5$ performs best on both LongMemEval and LoCoMo. 
A smaller threshold tends to under-truncate and introduce noisy evidence, while a larger threshold may stop expansion too early and miss necessary intermediate evidence. 
The consistent trend across two benchmarks suggests that CoM does not require dataset-specific tuning of $\beta$.

\begin{table}[t]
    \small
    \centering
    \renewcommand{\arraystretch}{1.2} 
    
    \setlength{\tabcolsep}{9.5pt} 
    
    \begin{tabular}{lccc}
        \toprule
        \textbf{Dataset} & \textbf{$\beta=0.3$} & \textbf{$\beta=0.5$} & \textbf{$\beta=0.7$} \\
        \midrule
        LongMemEval & 73.95 & \textbf{76.40} & 70.14 \\
        LoCoMo & 67.52 & \textbf{70.97} & 66.48 \\
        \bottomrule
    \end{tabular}
    
    \caption{Analysis of the adaptive truncation threshold $\beta$ using Qwen3-32B. Accuracy is reported in percentage.}
    \label{tab:beta_sensitivity}
\end{table}

\paragraph{Robustness across embedding models.}
To examine whether CoM depends on a particular embedding model, we replace Qwen3-Embedding-8B with \texttt{text-embedding-3-small}, keeping Qwen3-32B as the backbone LLM. As shown in Appendix~\ref{app:embedding_robustness}, CoM consistently outperforms RAG on both LongMemEval and LoCoMo. This suggests that CoM's gains do not rely on a specific embedding backbone, but stem from organizing retrieved fragments into context-aware memory chains.

\subsection{Upper Bound Analysis}
\label{sec:upper_bound_analysis}

To rigorously benchmark the theoretical limits and efficacy of our approach, we establish two distinct baselines: a state-of-the-art long-context model and an oracle setting for our backbone models.

\paragraph{Long-Context Baseline.} We employ \textit{Gemini-2.5-pro}, utilizing its 1M token context window to process the full input without truncation. Given that \textit{LongMemEval} samples average 105k tokens, this baseline effectively bypasses retrieval-induced information loss, serving as a robust upper bound for lossless context understanding.

\paragraph{Oracle Setting (Evidence Only).} To isolate the intrinsic reasoning capabilities of our backbones (\textit{GPT-4o-mini} and \textit{Qwen3-32B}) from retrieval errors, we evaluate an Oracle setting. Here, models are fed only the manually annotated ground-truth evidence. This configuration eliminates noise, quantifying the performance ceiling achievable when perfect information is guaranteed.

\paragraph{Analysis and Conclusion.} As detailed in Table~\ref{tab:upper_bound}, \textit{Gemini-2.5-pro} achieves the highest accuracy (Total: 89.20\%), confirming the advantage of full-context processing. However, this precision entails steep computational costs (122.89k tokens, 8512.10s). The Oracle results demarcate the reasoning limits of the smaller backbones; our method approaches this ceiling (e.g., 76.40\% vs. 81.80\% for Qwen3), indicating that our approach successfully captures the majority of critical information. Crucially, while \textit{Gemini-2.5-pro} offers marginal accuracy gains, our method is significantly more efficient—reducing token consumption by $14\times$ and latency by $4\times$. This presents a superior trade-off for practical applications where the overhead of processing massive contexts is prohibitive.

\begin{table*}[t]
\centering
\resizebox{1.0\textwidth}{!}{%
\begin{tabular}{lcccc c r r} 
\toprule
\multirow{2}{*}{\textbf{Method}} & \multicolumn{4}{c}{\textbf{Sub-category Acc (\%)}} & \multicolumn{1}{c}{\textbf{Total}} & \multicolumn{1}{c}{\textbf{Tokens}} & \multicolumn{1}{c}{\textbf{RunTime}} \\
\cmidrule(lr){2-5} 
 & \textbf{Single-hop} & \textbf{Multi-hop} & \textbf{Temporal} & \textbf{Knowledge} & \textbf{Acc (\%)} & \textbf{(k)} & \textbf{(s)} \\
\midrule
Full Context (Gemini-2.5-pro) & \textbf{90.38} & \textbf{86.47} & \textbf{90.98} & \textbf{88.46} & \textbf{89.20} & 122.89 & 8512.10 \\
\midrule
Evidence Only (GPT-4o-mini) & 83.33 & 82.71 & 69.17 & \textbf{88.46} & 80.20 & \textbf{0.30} & 668.36 \\
\rowcolor{gray_bg} Ours (GPT-4o-mini) & 84.62 & 65.41 & 65.41 & 83.33 & 74.20 & 8.24 & 2730.03 \\
\midrule
Evidence Only (Qwen3-32B) & 89.74 & 81.20 & 70.68 & 85.90 & 81.80 & 0.35 & \textbf{456.20} \\
\rowcolor{gray_bg} Ours (Qwen3-32B) & 86.54 & 69.92 & 69.17 & 79.49 & 76.40 & 8.81 & 2002.45 \\
\bottomrule
\end{tabular}%
}
\caption{Performance comparison between Full-Context SOTA (Gemini-2.5-pro), Evidence-Only (with only the ground-truth target information required to answer the question), and our method on LongMemEval dataset. }
\label{tab:upper_bound}
\end{table*}

\subsection{Error Analysis}
\label{sec:error_analysis}


\begin{figure}[t]  
    \centering
    \includegraphics[width=\columnwidth]{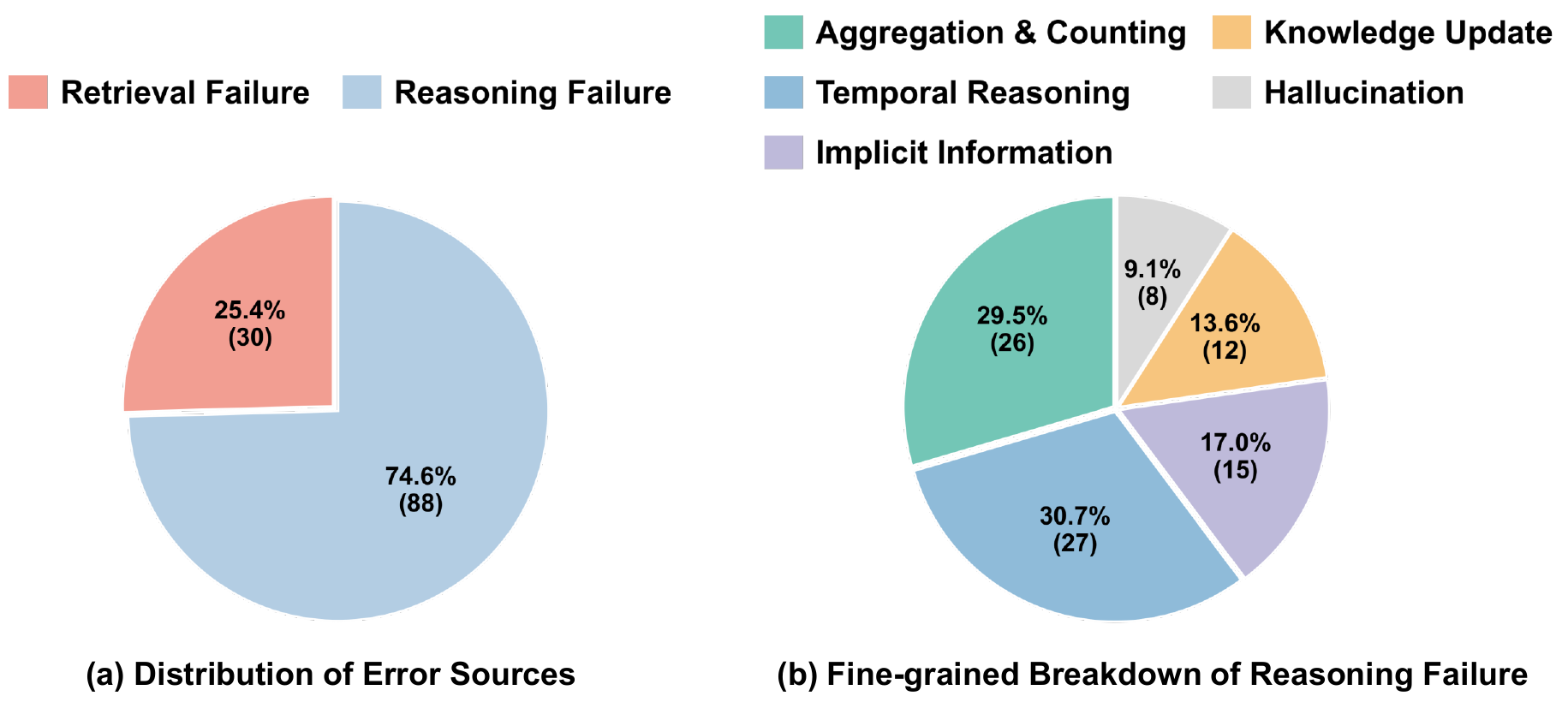}
    \caption{Distribution of error types on the LongMemEval dataset. Reasoning failures constitute the majority of errors compared to retrieval failure.}
    \label{fig:error_breakdown}
\end{figure}

To systematically diagnose failure modes, we conducted a comprehensive analysis of 118 error samples from the LongMemEval dataset. As illustrated in Figure~\ref{fig:error_breakdown}(a), reasoning failures constitute the vast majority of errors (74.6\%), whereas retrieval failures account for a relatively minor portion (25.4\%). This distribution suggests that the primary performance bottleneck lies in the model's capability to interpret and reason over context, rather than the inability to access relevant information.

\paragraph{Retrieval Failure}This category accounts for a minority of errors. Primary causes for retrieval failure include information dispersion in multi-hop questions, middle-position information loss in long contexts, and the semantic misalignment often encountered between queries and source documents.

\paragraph{Reasoning Failure}
Reasoning errors dominate the failure modes, persisting even after successful evidence retrieval. In these cases, the model accesses the context but fails to effectively extract or interpret the information. We categorize these failures into five subtypes:
(1) \textbf{Temporal Reasoning (30.7\%):} Models struggle to reconstruct accurate chronological timelines, frequently misinterpreting relative expressions or misordering sequential events;
(2) \textbf{Aggregation \& Counting (29.5\%):} Prominent in questions requiring statistical synthesis, this error arises when models fail to distinguish between entities actively \textit{acquired} and those merely \textit{mentioned}, resulting in systematic over-counting;
(3) \textbf{Implicit Information (17.0\%):} Models lack the sensitivity to infer latent user intents or constraints from behavioral patterns, consequently generating generic responses that lack necessary personalization;
(4) \textbf{Knowledge Update (13.6\%):} Models exhibit inertia in tracking dynamic state changes, often prioritizing semantically relevant but outdated information over the most recent evidence; and
(5) \textbf{Hallucination (9.1\%):} Occurring primarily in abstention tasks, models fabricate details based on tangential associations instead of correctly acknowledging the lack of evidence.

\section{Conclusion}

This work shows that effective agent memory can be achieved without increasingly complex memory construction. We propose \ours, a Chain-of-Memory framework that shifts the focus to lightweight retrieval and principled memory utilization by organizing retrieved fragments into dynamic inference paths and pruning distracting contexts through adaptive truncation. Experiments on LongMemEval and LoCoMo show that \ours improves reasoning accuracy, especially on multi-hop and temporal questions, while substantially reducing token consumption and runtime. These results demonstrate that structured memory utilization can bridge the gap between retrieved evidence and final reasoning accuracy. Rather than relying on heavier memory structures, \ours improves how retrieved fragments are connected, filtered, and presented to the LLM. Overall, our findings highlight lightweight construction, context-aware memory organization, and system-level efficiency as practical directions for building efficient long-term memory systems for LLM agents.

\section*{Limitations}

Despite its effectiveness, \ours still has several limitations.
First, \ours relies on embedding-based semantic similarity to retrieve and evolve memory chains. Its performance may therefore degrade when the embedding model fails to capture fine-grained temporal, pragmatic, or implicit relations between memory fragments.
Second, although adaptive truncation helps reduce noise and redundancy, it may discard subtle evidence in cases that require unusually long or weakly connected reasoning paths.
Third, our experiments focus on textual long-term memory benchmarks, leaving multi-modal memories, such as vision-language interaction histories and tool-use traces, for future work.
Finally, real-world agents may involve more open-ended goals, evolving user preferences, and noisier memory updates than current benchmarks cover. Further validation in interactive deployment settings is needed to assess the practical robustness of \ours under such conditions.

\section*{Acknowledgments}

We would like to thank the Director's Fund Project of State Key Laboratory of AI Safety for their support. This work was also supported in part by the Strategic Priority Research Program of the CAS (No.~XDB0680302) and the Young Elite Scientists Sponsorship Program of the Beijing High Innovation Plan (No.~20250924).

\bibliography{custom}

@article{qwen3embedding,
  title={Qwen3 Embedding: Advancing Text Embedding and Reranking Through Foundation Models},
  author={Zhang, Yanzhao and Li, Mingxin and Long, Dingkun and Zhang, Xin and Lin, Huan and Yang, Baosong and Xie, Pengjun and Yang, An and Liu, Dayiheng and Lin, Junyang and Huang, Fei and Zhou, Jingren},
  journal={arXiv preprint arXiv:2506.05176},
  year={2025}
}

@misc{qwen3technicalreport,
      title={Qwen3 Technical Report}, 
      author={Qwen Team},
      year={2025},
      eprint={2505.09388},
      archivePrefix={arXiv},
      primaryClass={cs.CL},
      url={https://arxiv.org/abs/2505.09388}, 
}

@article{wu2024longmemeval,
  title={Longmemeval: Benchmarking chat assistants on long-term interactive memory},
  author={Wu, Di and Wang, Hongwei and Yu, Wenhao and Zhang, Yuwei and Chang, Kai-Wei and Yu, Dong},
  journal={arXiv preprint arXiv:2410.10813},
  year={2024}
}

@article{maharana2024evaluating,
  title={Evaluating very long-term conversational memory of llm agents},
  author={Maharana, Adyasha and Lee, Dong-Ho and Tulyakov, Sergey and Bansal, Mohit and Barbieri, Francesco and Fang, Yuwei},
  journal={arXiv preprint arXiv:2402.17753},
  year={2024}
}

@article{chhikara2025mem0,
  title={Mem0: Building production-ready ai agents with scalable long-term memory},
  author={Chhikara, Prateek and Khant, Dev and Aryan, Saket and Singh, Taranjeet and Yadav, Deshraj},
  journal={arXiv preprint arXiv:2504.19413},
  year={2025}
}

@article{xu2025mem,
  title={A-mem: Agentic memory for llm agents},
  author={Xu, Wujiang and Liang, Zujie and Mei, Kai and Gao, Hang and Tan, Juntao and Zhang, Yongfeng},
  journal={arXiv preprint arXiv:2502.12110},
  year={2025}
}

@article{gao2023retrieval,
  title={Retrieval-augmented generation for large language models: A survey},
  author={Gao, Yunfan and Xiong, Yun and Gao, Xinyu and Jia, Kangxiang and Pan, Jinliu and Bi, Yuxi and Dai, Yixin and Sun, Jiawei and Wang, Haofen and Wang, Haofen},
  journal={arXiv preprint arXiv:2312.10997},
  volume={2},
  number={1},
  year={2023}
}

@article{hurst2024gpt,
  title={Gpt-4o system card},
  author={Hurst, Aaron and Lerer, Adam and Goucher, Adam P and Perelman, Adam and Ramesh, Aditya and Clark, Aidan and Ostrow, AJ and Welihinda, Akila and Hayes, Alan and Radford, Alec and others},
  journal={arXiv preprint arXiv:2410.21276},
  year={2024}
}

@inproceedings{sarthi2024raptor,
  title={Raptor: Recursive abstractive processing for tree-organized retrieval},
  author={Sarthi, Parth and Abdullah, Salman and Tuli, Aditi and Khanna, Shubh and Goldie, Anna and Manning, Christopher D},
  booktitle={The Twelfth International Conference on Learning Representations},
  year={2024}
}

@inproceedings{liu2025passage,
  title={Passage segmentation of documents for extractive question answering},
  author={Liu, Zuhong and Simon, Charles-Elie and Caspani, Fabien},
  booktitle={European Conference on Information Retrieval},
  pages={345--352},
  year={2025},
  organization={Springer}
}

@inproceedings{qu2025semantic,
  title={Is semantic chunking worth the computational cost?},
  author={Qu, Renyi and Tu, Ruixuan and Bao, Forrest},
  booktitle={Findings of the Association for Computational Linguistics: NAACL 2025},
  pages={2155--2177},
  year={2025}
}

@inproceedings{pan2025secom,
  title={Secom: On memory construction and retrieval for personalized conversational agents},
  author={Pan, Zhuoshi and Wu, Qianhui and Jiang, Huiqiang and Luo, Xufang and Cheng, Hao and Li, Dongsheng and Yang, Yuqing and Lin, Chin-Yew and Zhao, H Vicky and Qiu, Lili and others},
  booktitle={The Thirteenth International Conference on Learning Representations},
  year={2025}
}

@inproceedings{duarte2024lumberchunker,
  title={Lumberchunker: Long-form narrative document segmentation},
  author={Duarte, Andr{\'e} V and Marques, Jo{\~a}o DS and Gra{\c{c}}a, Miguel and Freire, Miguel and Li, Lei and Oliveira, Arlindo L},
  booktitle={Findings of the Association for Computational Linguistics: EMNLP 2024},
  pages={6473--6486},
  year={2024}
}

@article{zhao2024meta,
  title={Meta-Chunking: Learning Text Segmentation and Semantic Completion via Logical Perception},
  author={Zhao, Jihao and Ji, Zhiyuan and Feng, Yuchen and Qi, Pengnian and Niu, Simin and Tang, Bo and Xiong, Feiyu and Li, Zhiyu},
  journal={arXiv preprint arXiv:2410.12788},
  year={2024}
}

@inproceedings{yu2024chain,
  title={Chain-of-note: Enhancing robustness in retrieval-augmented language models},
  author={Yu, Wenhao and Zhang, Hongming and Pan, Xiaoman and Cao, Peixin and Ma, Kaixin and Li, Jian and Wang, Hongwei and Yu, Dong},
  booktitle={Proceedings of the 2024 conference on empirical methods in natural language processing},
  pages={14672--14685},
  year={2024}
}

@article{wang2025inducing,
  title={Inducing programmatic skills for agentic tasks},
  author={Wang, Zora Zhiruo and Gandhi, Apurva and Neubig, Graham and Fried, Daniel},
  journal={arXiv preprint arXiv:2504.06821},
  year={2025}
}

@inproceedings{zhao2024expel,
  title={Expel: Llm agents are experiential learners},
  author={Zhao, Andrew and Huang, Daniel and Xu, Quentin and Lin, Matthieu and Liu, Yong-Jin and Huang, Gao},
  booktitle={Proceedings of the AAAI Conference on Artificial Intelligence},
  volume={38},
  number={17},
  pages={19632--19642},
  year={2024}
}

@inproceedings{sun2025rearter,
  title={Rearter: Retrieval-augmented reasoning with trustworthy process rewarding},
  author={Sun, Zhongxiang and Wang, Qipeng and Yu, Weijie and Zang, Xiaoxue and Zheng, Kai and Xu, Jun and Zhang, Xiao and Song, Yang and Li, Han},
  booktitle={Proceedings of the 48th International ACM SIGIR Conference on Research and Development in Information Retrieval},
  pages={1251--1261},
  year={2025}
}

@inproceedings{zhong2024memorybank,
  title={Memorybank: Enhancing large language models with long-term memory},
  author={Zhong, Wanjun and Guo, Lianghong and Gao, Qiqi and Ye, He and Wang, Yanlin},
  booktitle={Proceedings of the AAAI Conference on Artificial Intelligence},
  volume={38},
  number={17},
  pages={19724--19731},
  year={2024}
}

@article{liu2025advances,
  title={Advances and challenges in foundation agents: From brain-inspired intelligence to evolutionary, collaborative, and safe systems},
  author={Liu, Bang and Li, Xinfeng and Zhang, Jiayi and Wang, Jinlin and He, Tanjin and Hong, Sirui and Liu, Hongzhang and Zhang, Shaokun and Song, Kaitao and Zhu, Kunlun and others},
  journal={arXiv preprint arXiv:2504.01990},
  year={2025}
}

@article{mei2025survey,
  title={A survey of context engineering for large language models},
  author={Mei, Lingrui and Yao, Jiayu and Ge, Yuyao and Wang, Yiwei and Bi, Baolong and Cai, Yujun and Liu, Jiazhi and Li, Mingyu and Li, Zhong-Zhi and Zhang, Duzhen and others},
  journal={arXiv preprint arXiv:2507.13334},
  year={2025}
}

@article{rasmussen2025zep,
  title={Zep: a temporal knowledge graph architecture for agent memory},
  author={Rasmussen, Preston and Paliychuk, Pavlo and Beauvais, Travis and Ryan, Jack and Chalef, Daniel},
  journal={arXiv preprint arXiv:2501.13956},
  year={2025}
}

@article{guo2024lightrag,
  title={Lightrag: Simple and fast retrieval-augmented generation},
  author={Guo, Zirui and Xia, Lianghao and Yu, Yanhua and Ao, Tu and Huang, Chao},
  journal={arXiv preprint arXiv:2410.05779},
  year={2024}
}

@article{dong2025youtu,
  title={Youtu-graphrag: Vertically unified agents for graph retrieval-augmented complex reasoning},
  author={Dong, Junnan and An, Siyu and Yu, Yifei and Zhang, Qian-Wen and Luo, Linhao and Huang, Xiao and Wu, Yunsheng and Yin, Di and Sun, Xing},
  journal={arXiv preprint arXiv:2508.19855},
  year={2025}
}

@article{packer2023memgpt,
  title={MemGPT: Towards LLMs as Operating Systems.},
  author={Packer, Charles and Fang, Vivian and Patil, Shishir\_G and Lin, Kevin and Wooders, Sarah and Gonzalez, Joseph\_E},
  year={2023},
  publisher={ArXiv}
}

@article{wang2023enhancing,
  title={Enhancing large language model with self-controlled memory framework},
  author={Wang, Bing and Liang, Xinnian and Yang, Jian and Huang, Hui and Wu, Shuangzhi and Wu, Peihao and Lu, Lu and Ma, Zejun and Li, Zhoujun},
  journal={arXiv preprint arXiv:2304.13343},
  year={2023}
}

@article{rezazadeh2024isolated,
  title={From isolated conversations to hierarchical schemas: Dynamic tree memory representation for llms},
  author={Rezazadeh, Alireza and Li, Zichao and Wei, Wei and Bao, Yujia},
  journal={arXiv preprint arXiv:2410.14052},
  year={2024}
}

@article{yang2024swe,
  title={Swe-agent: Agent-computer interfaces enable automated software engineering},
  author={Yang, John and Jimenez, Carlos E and Wettig, Alexander and Lieret, Kilian and Yao, Shunyu and Narasimhan, Karthik and Press, Ofir},
  journal={Advances in Neural Information Processing Systems},
  volume={37},
  pages={50528--50652},
  year={2024}
}

@article{singh2025agentic,
  title={Agentic reasoning and tool integration for llms via reinforcement learning},
  author={Singh, Joykirat and Magazine, Raghav and Pandya, Yash and Nambi, Akshay},
  journal={arXiv preprint arXiv:2505.01441},
  year={2025}
}

@article{zhang2025landscape,
  title={The landscape of agentic reinforcement learning for llms: A survey},
  author={Zhang, Guibin and Geng, Hejia and Yu, Xiaohang and Yin, Zhenfei and Zhang, Zaibin and Tan, Zelin and Zhou, Heng and Li, Zhongzhi and Xue, Xiangyuan and Li, Yijiang and others},
  journal={arXiv preprint arXiv:2509.02547},
  year={2025}
}

@inproceedings{fei2024extending,
  title={Extending context window of large language models via semantic compression},
  author={Fei, Weizhi and Niu, Xueyan and Zhou, Pingyi and Hou, Lu and Bai, Bo and Deng, Lei and Han, Wei},
  booktitle={Findings of the Association for Computational Linguistics: ACL 2024},
  pages={5169--5181},
  year={2024}
}

@article{zhang2025survey,
  title={A survey on the memory mechanism of large language model-based agents},
  author={Zhang, Zeyu and Dai, Quanyu and Bo, Xiaohe and Ma, Chen and Li, Rui and Chen, Xu and Zhu, Jieming and Dong, Zhenhua and Wen, Ji-Rong},
  journal={ACM Transactions on Information Systems},
  volume={43},
  number={6},
  pages={1--47},
  year={2025},
  publisher={ACM New York, NY}
}

@inproceedings{wang2024adapting,
  title={Adapting llms for efficient context processing through soft prompt compression},
  author={Wang, Cangqing and Yang, Yutian and Li, Ruisi and Sun, Dan and Cai, Ruicong and Zhang, Yuzhu and Fu, Chengqian},
  booktitle={Proceedings of the International Conference on Modeling, Natural Language Processing and Machine Learning},
  pages={91--97},
  year={2024}
}

@article{liu2024lost,
  title={Lost in the middle: How language models use long contexts},
  author={Liu, Nelson F and Lin, Kevin and Hewitt, John and Paranjape, Ashwin and Bevilacqua, Michele and Petroni, Fabio and Liang, Percy},
  journal={Transactions of the Association for Computational Linguistics},
  volume={12},
  pages={157--173},
  year={2024}
}

@article{li2025memos,
  title={MemOS: An Operating System for Memory-Augmented Generation (MAG) in Large Language Models},
  author={Li, Zhiyu and Song, Shichao and Wang, Hanyu and Niu, Simin and Chen, Ding and Yang, Jiawei and Xi, Chenyang and Lai, Huayi and Zhao, Jihao and Wang, Yezhaohui and others},
  journal={arXiv preprint arXiv:2505.22101},
  year={2025}
}

@software{Chase_LangChain_2022, author = {Chase, Harrison}, month = oct, title = {{LangChain}}, url = {https://github.com/hwchase17/langchain}, year = {2022} }

\clearpage
\appendix
\onecolumn

\section{Appendix}

\subsection{Prompts}
\label{sec:appendix_prompts}

\noindent
\begin{UnifiedPromptBox}{LongMemEval Answer Prompt}
    \textbf{System Prompt:} \\
    You are a helpful expert assistant answering questions from the user based on the provided context.

    \vspace{2mm}
    \textbf{User Prompt:} \\
    Your task is to briefly answer the question. You are given the following context from the previous conversation. If you don't know how to answer the question, abstain from answering.

    \vspace{2mm}
    \textbf{Context:} \texttt{\{context\}} \\
    \textbf{Question:} \texttt{\{question\}}
\end{UnifiedPromptBox}

\begin{UnifiedPromptBox}{LoCoMo Answer Prompt}
    \textbf{System Prompt:} \\
    You are a \textbf{Memory-Based Question Answering Agent}. Answer questions based on the provided memories.

    \vspace{2mm}
    \textbf{\texttt{<Instructions>}}
    \begin{enumerate}[leftmargin=*, nosep, label=\arabic*.]
        \item \textbf{For time-related questions (with words like ``when''):}
        \begin{itemize}[leftmargin=3mm, topsep=2pt, itemsep=1pt]
            \item \textbf{Step 1:} Identify the Timestamp of the memory.
            \item \textbf{Step 2:} Identify any relative time reference.
            \item \textbf{Step 3:} Calculate the \textbf{EXACT ABSOLUTE} date.
            \item \textit{Example 1:} [Timestamp: 4 May 2022] + ``last year'' $\to$ Answer: ``last year before 4 May 2022''
        \end{itemize}

        \vspace{2pt}
        \item \textbf{Extract specific details:} Use exact names, dates, places from memories.
        \vspace{2pt}
        \item \textbf{Be comprehensive for list questions:} List ALL activities found.
        \vspace{2pt}
        \item \textbf{Contradictions:} If memories conflict, use the most recent one.
    \end{enumerate}
    \textbf{\texttt{</Instructions>}}

    \PromptRule

    \textbf{User Prompt:} \\
    \texttt{<Memories>} \\
    \texttt{\{memories\}} \\
    \texttt{</Memories>}

    \vspace{2mm}
    \texttt{<Question>\{question\}</Question>}

    \vspace{2mm}
    \texttt{Answer:}
\end{UnifiedPromptBox}

\begin{UnifiedPromptBox}{LongMemEval Judge Prompt}
    You are an expert judge evaluating whether a model's response correctly answers a question according to the reference answer or rubric.

    \vspace{3mm}
    \textbf{1. Basic Evaluation} \\
    I will give you a question, a correct answer, and a response from a model. Please answer yes if the response contains the correct answer.

    \vspace{2mm}
    \textbf{Question:} \texttt{\{\}} \\
    \textbf{Correct Answer:} \texttt{\{\}} \\
    \textbf{Model Response:} \texttt{\{\}}

    \vspace{2mm}
    \noindent Is the model response correct? Answer yes or no only.

    \PromptRule

    \textbf{2. Temporal Reasoning} \\
    I will give you a question, a correct answer, and a response from a model. If the response is equivalent or contains all intermediate steps, answer yes. Do not penalize off-by-one errors for the number of days.

    \vspace{2mm}
    \textbf{Question:} \texttt{\{\}} \\
    \textbf{Correct Answer:} \texttt{\{\}} \\
    \textbf{Model Response:} \texttt{\{\}}

    \vspace{2mm}
    \noindent Is the model response correct? Answer yes or no only.

    \PromptRule

    \textbf{3. Knowledge Update} \\
    If the response contains some previous information along with an updated answer, the response should be considered correct.

    \vspace{2mm}
    \textbf{Question:} \texttt{\{\}} \\
    \textbf{Correct Answer:} \texttt{\{\}} \\
    \textbf{Model Response:} \texttt{\{\}}

    \vspace{2mm}
    \noindent Is the model response correct? Answer yes or no only.

    \PromptRule

    \textbf{4. User Preference} \\
    Please answer yes if the response satisfies the desired response based on the rubric.

    \vspace{2mm}
    \textbf{Question:} \texttt{\{\}} \\
    \textbf{Rubric:} \texttt{\{\}} \\
    \textbf{Model Response:} \texttt{\{\}}

    \vspace{2mm}
    \noindent Is the model response correct? Answer yes or no only.

    \PromptRule

    \textbf{5. Abstention (Unanswerable)} \\
    Please answer yes if the model correctly identifies the question as unanswerable.

    \vspace{2mm}
    \textbf{Question:} \texttt{\{\}} \\
    \textbf{Explanation:} \texttt{\{\}} \\
    \textbf{Model Response:} \texttt{\{\}}

    \vspace{2mm}
    \noindent Does the model correctly identify the question as unanswerable? Answer yes or no only.
\end{UnifiedPromptBox}

\begin{UnifiedPromptBox}{LoCoMo Judge Prompt}
    You are an expert judge evaluating whether a model's prediction correctly answers a question compared to the reference answer.

    \vspace{3mm}
    \textbf{Question:} \texttt{\{\}} \\
    \textbf{Reference Answer:} \texttt{\{\}} \\
    \textbf{Model Prediction:} \texttt{\{\}}

    \vspace{3mm}
    \noindent Your task is to determine if the model's prediction is semantically equivalent to the reference answer. Consider the following:
    \begin{enumerate}[leftmargin=*, nosep, label=\arabic*.]
        \item The prediction may be phrased differently but convey the same meaning.
        \item Minor differences in wording are acceptable if the core information matches.
        \item For dates, consider different formats as equivalent (e.g., ``7 May 2023'' vs ``May 7, 2023'').
        \item For numbers, consider ``2022'' vs ``Last year'' as potentially equivalent depending on context.
        \item For descriptive answers, check if the key information is present.
    \end{enumerate}

    \vspace{3mm}
    \noindent Respond with ONLY ONE WORD:
    \begin{itemize}[leftmargin=8mm, nosep, label={-}]
        \item ``CORRECT'' if the prediction matches the reference answer.
        \item ``INCORRECT'' if the prediction does not match the reference answer.
    \end{itemize}

    \vspace{3mm}
    \noindent \textbf{Your response:}
\end{UnifiedPromptBox}

\clearpage
\twocolumn

\subsection{Comparison with Re-ranking and Iterative Retrieval}
\label{app:reranking_comparison}

CoM is related to post-retrieval re-ranking and iterative retrieval, but differs in both objective and operation. Re-ranking methods typically assign each retrieved segment an independent relevance score with respect to the query, and then select a reordered top-$k$ set. Iterative retrieval methods often expand or reformulate the retrieval query over multiple rounds to retrieve additional evidence. In contrast, CoM keeps memory construction and initial retrieval lightweight, and focuses on organizing the retrieved fragments into chain-structured inference paths. Each expansion step is conditioned not only on the original query but also on the evolving chain context, allowing CoM to recover multi-hop and temporal dependencies that may not be captured by global relevance alone.

To test whether stronger post-retrieval scoring can explain CoM's gains, we compare with a cross-encoder re-ranking baseline. This baseline first retrieves an expanded candidate pool and then uses \texttt{bge-reranker-base} to select the final top-$k$ segments for generation. As shown in Table~\ref{tab:rerank_longmemeval} and Table~\ref{tab:rerank_locomo}, even with a dedicated re-ranker, this baseline remains below CoM on both benchmarks. This suggests that CoM's improvements do not merely come from assigning better independent relevance scores, but from dynamically organizing retrieved fragments into context-aware memory chains.

\begin{table}[t]
    \small
    \centering
    \resizebox{\columnwidth}{!}{
        \begin{tabular}{lccc}
            \toprule
            \textbf{Method} & \textbf{Acc.} & \textbf{Tokens} & \textbf{Time} \\
            \midrule
            Rerank (GPT-4o-mini) & 62.73 & 11.01 & 3967.52 \\
            CoM (GPT-4o-mini) & \textbf{74.20} & \textbf{8.24} & \textbf{2730.03} \\
            Rerank (Qwen3-32B) & 63.20 & 10.95 & 2702.47 \\
            CoM (Qwen3-32B) & \textbf{76.40} & \textbf{8.81} & \textbf{2002.45} \\
            \bottomrule
        \end{tabular}
    }
    \caption{Comparison with a cross-encoder re-ranking baseline on LongMemEval. Tokens are reported in thousands.}
    \label{tab:rerank_longmemeval}
\end{table}

\begin{table}[t]
    \small
    \centering
    \resizebox{\columnwidth}{!}{
        \begin{tabular}{lccc}
            \toprule
            \textbf{Method} & \textbf{Acc.} & \textbf{Tokens} & \textbf{Time} \\
            \midrule
            Rerank (GPT-4o-mini) & 68.31 & \textbf{3.53} & 3413.43 \\
            CoM (GPT-4o-mini) & \textbf{72.86} & 4.35 & \textbf{3057.50} \\
            Rerank (Qwen3-32B) & 65.32 & \textbf{3.69} & 2805.66 \\
            CoM (Qwen3-32B) & \textbf{70.97} & 4.77 & \textbf{2050.48} \\
            \bottomrule
        \end{tabular}
    }
    \caption{Comparison with a cross-encoder re-ranking baseline on LoCoMo. Tokens are reported in thousands.}
    \label{tab:rerank_locomo}
\end{table}

\subsection{Robustness Across Embedding Models}
\label{app:embedding_robustness}

To evaluate whether CoM depends on a specific embedding model, we replace Qwen3-Embedding-8B with \texttt{text-embedding-3-small} and compare CoM with RAG using Qwen3-32B as the backbone LLM. Tables~\ref{tab:embedding_longmemeval} and~\ref{tab:embedding_locomo} show that CoM consistently improves overall accuracy on both benchmarks.

\begin{table}[t]
    \small
    \centering
    \resizebox{\columnwidth}{!}{
        \begin{tabular}{lccccc}
            \toprule
            \textbf{Method} & \textbf{S-hop} & \textbf{M-hop} & \textbf{Temp} & \textbf{Kno} & \textbf{Overall} \\
            \midrule
            RAG & 80.77 & 54.14 & 53.38 & 74.36 & 65.40 \\
            CoM & \textbf{84.62} & \textbf{68.42} & \textbf{67.67} & \textbf{79.49} & \textbf{75.00} \\
            \bottomrule
        \end{tabular}
    }
    \caption{Performance on LongMemEval using \texttt{text-embedding-3-small} with Qwen3-32B.}
    \label{tab:embedding_longmemeval}
\end{table}

\begin{table}[t]
    \small
    \centering
    \resizebox{\columnwidth}{!}{
        \begin{tabular}{lccccc}
            \toprule
            \textbf{Method} & \textbf{S-hop} & \textbf{M-hop} & \textbf{Temp} & \textbf{Kno} & \textbf{Overall} \\
            \midrule
            RAG & 74.31 & 43.26 & 66.35 & 43.75 & 65.06 \\
            CoM & \textbf{81.92} & \textbf{43.61} & \textbf{69.47} & \textbf{46.87} & \textbf{70.12} \\
            \bottomrule
        \end{tabular}
    }
    \caption{Performance on LoCoMo using \texttt{text-embedding-3-small} with Qwen3-32B.}
    \label{tab:embedding_locomo}
\end{table}

\end{document}